\documentclass{article}
\usepackage{spconf,amsmath,graphicx}
\usepackage[utf8]{inputenc}
\usepackage{booktabs}
\usepackage{amsfonts,amssymb,amsthm}
\usepackage{algorithm}
\usepackage{algpseudocode}
\usepackage{multirow}
\usepackage{xspace,xcolor}
\usepackage{microtype}
\usepackage{nicefrac}
\usepackage{marvosym}
\usepackage{enumitem}
\usepackage{bm}
\usepackage{cite}
\usepackage{balance}
\usepackage{adjustbox}
\usepackage[colorlinks,linkcolor=blue, citecolor=blue]{hyperref}  
\usepackage{orcidlink}
\usepackage{subcaption} 
\usepackage{float}     

\usepackage{booktabs}
\usepackage{multirow}
\usepackage{adjustbox}
\usepackage[table]{xcolor}
\definecolor{headergray}{gray}{0.93}
\definecolor{oursgray}{gray}{0.96}

\allowdisplaybreaks
\title{Simulation-Free Learning of GP-SDEs from Irregular Observations}

\name{
Zhidi Lin$^{\star}$, 
Yuhao Liu$^{\sharp}$, 
Ying Li$^{\dagger}$, 
Edwin Fong$^{\star}$, 
Petar Djuri\'c$^{\diamond}$
}
\address{
$\star$ The University of Hong Kong, \,$\sharp$ Capital One, 
\\
$\dagger$ National University of Singapore, \,
$\diamond$ Stony Brook University
}

\begin{document}
\ninept
\maketitle
\begin{abstract}
Gaussian process stochastic differential equations (GP-SDEs) provide a flexible Bayesian model for unknown continuous-time state dynamics with uncertainty quantification, but learning and inference from noisy and irregular observations remain computationally challenging.
To address this issue, we propose GP-SDE Matching, a simulation-free variational framework for Bayesian GP drift learning and continuous-time state smoothing.
We analytically marginalize the sparse GP posterior to derive a tractable drift-matching objective that accounts for both the posterior mean and uncertainty of the unknown drift.
To handle irregular observations, we further introduce an irregular-time-aware variational state posterior that incorporates the actual observation times during both encoding and continuous-time marginal querying.
Experiments on the stochastic Lorenz--63 system demonstrate substantially improved drift recovery and state reconstruction under irregular observations, while five system identification benchmarks show robust forecasting under increasing observation sparsity and competitive performance against existing latent-SDE and state-space methods.
\end{abstract}


\vspace{-.03in}
\section{Introduction}
\label{sec:intro}
\vspace{-.07in}

Learning an unknown dynamical system from noisy and irregular measurements requires reasoning about uncertainty in both the latent state trajectory and the underlying evolution law. 
Gaussian process state-space models (GPSSMs) address this problem by placing a GP prior over the discrete-time transition function \cite{waxman2026sequential}, with inference approaches spanning particle methods \cite{frigola2013bayesian,liu2021gaussian}, variational inference \cite{frigola2014variational,doerr2018probabilistic,ialongo2019overcoming}, recognition-model smoothers \cite{eleftheriadis2017identification}, and online inference \cite{lin2023ensemble,liu2023sequential,lin2023towards_efficient,lin2025efficient}. 
However, standard GPSSMs evolve over a specified discrete step; unequal observation gaps therefore require modifying or repeatedly applying the transition over a chosen grid \cite{li2026gaussian}, increasing computational cost and discretization sensitivity.
Continuous-time models provide an alternative through Gaussian process stochastic differential equations (GP-SDEs) \cite{duncker2019learning}, which place a GP prior on the drift of an SDE.
Learning and inference in GP-SDEs remain challenging because nonlinear drift functions generally yield intractable latent state posteriors and transition densities. 

To address this challenge, existing GP-SDE methods approximate the latent state trajectory in different ways. Early work used approximate EM with locally linearized dynamics for sparse GP drift learning \cite{ruttor2013approximate}, while later approaches relied on simulated latent paths \cite{yildiz2018learning} or tractable variational diffusion processes \cite{archambeau2007gaussian,duncker2019learning} to enable joint state inference and Bayesian drift learning. 
Recent work has further explored structured GP dynamics and more efficient optimization \cite{hu2024modeling,hu2026sing}. Nevertheless, these methods generally require latent process simulation or temporal discretization during inference.
Accurately resolving the latent process often requires fine temporal grids, which increase computational and memory costs and make the results sensitive to the chosen discretization \cite{course2023amortized}.

More recently, simulation-free inference methods have been developed for latent SDEs by directly parameterizing continuous-time state marginals \cite{course2023amortized,bartosh2025sde}.  In particular, SDE Matching avoids latent process simulation by matching the drift induced by the variational state posterior to the generative SDE drift \cite{bartosh2025sde}. However, existing formulations mainly consider deterministic neural drifts, and their variational state parameterizations are designed for densely observed sequences rather than sparse and irregular observations.

To address these limitations, we develop GP-SDE Matching, extending simulation-free SDE Matching to Bayesian GP dynamics and irregularly sampled observations.
Our contributions are threefold:
\begin{itemize}[itemsep=2pt, topsep=2pt]
    \item [1)] Building on the continuous-time variational marginals of SDE Matching \cite{bartosh2025sde}, we introduce a sparse GP drift posterior that enables analytic evaluation of the GP expectation in the drift-matching objective. This allows joint Bayesian drift learning and state smoothing without latent SDE simulation, GP function sampling, or auxiliary solver grids.
    \item [2)] Unlike the uniformly sampled setting considered in \cite{bartosh2025sde}, we introduce an irregular-time-aware state posterior that incorporates the actual observation gaps during sequence encoding and the temporal distances to arbitrary query times, enabling inference from sparse and irregular observations.
    \item [3)] Experiments on the stochastic Lorenz--63 system show that the time-aware posterior substantially improves drift recovery and state reconstruction over the SDE Matching encoder under irregular observations. Across five system-identification benchmarks, GP-SDE Matching consistently achieves the best or second-best  predictive RMSE, while requiring lower training cost than grid-based SDE inference methods.
\end{itemize}
The remainder of this paper is organized as follows. Section \ref{sec:background} introduces the GP-SDE model and necessary background. Section \ref{sec:method} presents GP-SDE Matching. Section \ref{sec:experiment} reports experimental results, and Section \ref{sec:conclusion} concludes the paper.

\vspace{-.08in}
\section{Background}
\label{sec:background}
\vspace{-.08in}

\subsection{Gaussian processes (GPs)}
\label{sec:background_gp}
\vspace{-.05in}

A GP defines a distribution over functions, written as
$
f\!\sim\!\mathcal{GP}(m_f,k_f),
$
where $m_f$ and $k_f$ denote the mean and covariance functions, respectively \cite{williams2006gaussian}. For any finite collection of inputs, the corresponding function values are jointly Gaussian. Throughout, we assume a zero mean function and use a squared-exponential kernel with dimension-specific lengthscales.
For a vector-valued function
$\bm{f}:\mathbb{R}^{D}\rightarrow\mathbb{R}^{P}$, one convenient construction places an independent GP prior on each output,
$
f_p\!\sim\!\mathcal{GP}(m_p,k_p), p=1,\ldots,P.
$
More general output dependencies can be modeled using multi-output GPs \cite{alvarez2012kernels,lin2022output}.

For $B$ function evaluations, exact GP inference requires inverting a dense $B \!\times\! B$ kernel matrix and scales as $\mathcal{O}(B^3)$. Sparse variational GPs reduce this cost using $M\!\ll\! B$ inducing variables $\mathbf{u}=f(\mathbf{Z})$ at inducing inputs $\mathbf{Z}=\{\mathbf{z}_m\}_{m=1}^{M}$ \cite{titsias2009variational}. Under the GP prior,
$
\mathbf{u} \sim \mathcal{N}(\mathbf{m}_Z,\mathbf{K}_{ZZ}),
$
where $\mathbf{m}_Z=m_f(\mathbf{Z})$ and $\mathbf{K}_{ZZ}=k_f(\mathbf{Z},\mathbf{Z})$. Introducing a Gaussian variational posterior $q_{\lambda}(\mathbf{u})=\mathcal{N}(\mathbf{m}_u,\mathbf{S}_u)$ induces  an approximate GP posterior over the full function,
\begin{equation}
q_{\lambda}(f)=\int p(f\vert\mathbf{u})\, q_\lambda(\mathbf{u})\,\mathrm{d}\mathbf{u}
=\mathcal{GP}(\mu_{\lambda},\Sigma_{\lambda}),
\label{eq:VSGP}
\end{equation}
where $\lambda=\{\mathbf{m}_u,\mathbf{S}_u\}$ denotes the variational parameters, and
\begin{subequations}
\label{eq:VSGP_2}
\begin{align}
&\mu_{\lambda}(\mathbf{x}) =
m_f(\mathbf{x})
+\mathbf{k}_{xZ}\mathbf{K}_{ZZ}^{-1}
(\mathbf{m}_u-\mathbf{m}_Z),
\\
&\Sigma_{\lambda}(\mathbf{x},\mathbf{x}') =
k_f(\mathbf{x},\mathbf{x}')
+\mathbf{k}_{xZ}\mathbf{K}_{ZZ}^{-1}
(\mathbf{S}_u-\mathbf{K}_{ZZ})
\mathbf{K}_{ZZ}^{-1}\mathbf{k}_{Zx'},
\end{align}
\end{subequations}
with $\mathbf{k}_{xZ}=k_f(\mathbf{x},\mathbf{Z})$. This construction reduces the full-data computational cost to $\mathcal{O}(BM^2+M^3)$ while retaining a posterior distribution over the entire function. For notational simplicity, Eqs.~\eqref{eq:VSGP}-\eqref{eq:VSGP_2} here are written for a single output dimension and are applied independently to each component of vector-valued functions.

\subsection{GP-SDEs}
\label{sec:background_gpsde}

A GP-SDE places a GP prior on the unknown drift of an SDE.
Following \cite{duncker2019learning}, we consider
\begin{subequations}
\label{eq:gpsde_model}
\begin{alignat}{2}
&\mathbf{x}_0
\sim p(\mathbf{x}_0),
\qquad&
&\quad
\bm{f}
\sim \mathcal{GP}(\bm{m}_f,\bm{K}_f),
\\
&\mathrm{d}\mathbf{x}_t
=
\bm{f}(\mathbf{x}_t)\,\mathrm{d}t
+
\mathbf{G}\,\mathrm{d}\mathbf{W}_t,
&&
\\
&\mathbf{y}_n
=
\mathbf{g}(\mathbf{C}\mathbf{x}_{t_n})
+
\boldsymbol{\epsilon}_n,
\qquad&&\quad
\boldsymbol{\epsilon}_n
\sim
\mathcal{N}(\mathbf{0},\mathbf{R}_n).
\end{alignat}
\end{subequations}
Here $n=1,\ldots,N$, $\mathbf{x}_t\!\in\!\mathbb{R}^{D}$ is the latent state, and $\mathbf{W}_t$ is a standard Wiener process.
As in \cite{duncker2019learning}, we assume that the diffusion matrix $\mathbf{G}$ is fixed, known, and state-independent with $\mathbf{A}=\mathbf{G}\mathbf{G}^{\top}\succ 0$.
The drift may incorporate a known control input, $\bm{f}(\mathbf{x}_t,\mathbf{c}_t)$; we suppress $\mathbf{c}_t$ below to simplify notation.
The observations may occur at irregular times, $\mathcal{D} =
\{(t_n,\mathbf{y}_n)\}_{n=1}^{N},
0\leq t_1<\cdots<t_N\leq T,$
where $\mathbf{g}$ and $\mathbf{C}$ are known.

Let $\mathbf{X}=\{\mathbf{x}_t: 0\leq t\leq T\}$ denote the latent path, and let $\mathbb{P}_{\mathbf{f}}(\mathrm{d}\mathbf{X}\vert\mathbf{x}_0)$ denote the SDE path law induced by a realization of the drift $\mathbf{f}$. Introducing inducing variables
$\mathbf{u}=\mathbf{f}(\mathbf{Z})$, the augmented joint model factorizes as
\begin{align}
p(\mathcal{D},\mathbf{X},\mathbf{f},\mathbf{u})
\!=\!
p(\mathbf{u})
p(\mathbf{f}\vert\mathbf{u})
p(\mathbf{x}_0)
\mathbb{P}_{\mathbf{f}}
(\mathrm{d}\mathbf{X}\vert\mathbf{x}_0)
\prod_{n=1}^{N}
p(\mathbf{y}_n\vert\mathbf{x}_{t_n}).
\label{eq:gpsde_joint_model}
\end{align}
Learning and inference therefore require approximating the joint posterior over the latent path and GP drift while learning the associated model and kernel parameters. For nonlinear GP drifts, this posterior is analytically intractable because the induced SDE generally lacks closed-form transition densities. Existing methods consequently rely on approximate inference involving SDE simulation or temporal discretization, which can be  computationally demanding \cite{ruttor2013approximate,yildiz2018learning,duncker2019learning,hu2024modeling}.

To address this computational burden, Section~\ref{sec:method} develops GP-SDE Matching, a variational framework for joint Bayesian GP drift learning and continuous-time state smoothing that avoids latent SDE simulation, GP function sampling, and numerical solver grids.

\section{GP-SDE Matching}
\label{sec:method}

\subsection{Variational inference}
\label{sec:method_variational}
As noted above, exact posterior inference is intractable. We therefore introduce a tractable variational approximation and optimize the corresponding evidence lower bound (ELBO). Specifically, we use the following variational approximation:
\begin{equation}
q_{\lambda,\phi}(\mathbf{X},\mathbf{f},\mathbf{u}\mid\mathcal{D})
=
q_{\lambda}(\mathbf{u})\,
p(\mathbf{f}\vert\mathbf{u})\,
\mathbb{Q}_{\phi}(\mathrm{d}\mathbf{X}\vert\mathcal{D}),
\label{eq:joint_variational_gpsde}
\end{equation}
which imposes a mean-field assumption between the latent trajectory $\mathbf{X}$ and the GP variables $(\mathbf{f},\mathbf{u})$. Here $q_{\lambda}(\mathbf{u})$ is the sparse variational posterior over the inducing variables, while the conditional GP prior $p(\mathbf{f}\vert\mathbf{u})$ is retained exactly \cite{titsias2009variational}. The process $\mathbb{Q}_{\phi}$, parameterized by $\phi$, approximates the latent smoothing distribution. We denote its marginal at any $t\in[0,T]$ by $q_{\phi}(\mathbf{x}_t\vert\mathcal{D})$. 
The corresponding negative ELBO can be written as  \cite{bartosh2025sde,duncker2019learning}
\begin{equation}
-\log p(\mathcal{D})
\leq
\mathcal{L}
=
\mathcal{L}_{u}
+
\mathcal{L}_{0}
+
\mathcal{L}_{\mathrm{dyn}}
+
\mathcal{L}_{\mathrm{obs}},
\label{eq:gp_sde_simulation_free_nelbo}
\end{equation}
where $\mathcal{L}_{u} = \operatorname{KL}\!\left[ q_{\lambda}(\mathbf{u})
\,\|\,p(\mathbf{u})
\right],$ $\mathcal{L}_{0} = \operatorname{KL}\!\left[ q_{\phi}(\mathbf{x}_{0}\vert\mathcal{D})
\,\|\,p(\mathbf{x}_{0})
\right],$ and
\begin{align}
&\mathcal{L}_{\mathrm{obs}} = -\sum_{n=1}^{N} \mathbb{E}_{q_{\phi}(\mathbf{x}_{t_n}\vert\mathcal{D})} \left[ \log p(\mathbf{y}_{n}\vert\mathbf{x}_{t_n})\right].
\end{align}
Since $q_{\lambda}(\mathbf{u})=\mathcal{N}(\mathbf{m}_u,\mathbf{S}_u)$, the KL term $\mathcal{L}_u$ is available in closed form. Likewise, $\mathcal{L}_0$ is analytic when the initial distribution is Gaussian. The state marginals $q_{\phi}(\mathbf{x}_t\vert\mathcal{D})$ are parameterized by the irregular-time-aware variational family introduced in Section~\ref{sec:method_state_posterior}, and $\mathcal{L}_{\mathrm{obs}}$ is estimated using reparameterized samples from these marginals \cite{Kingma2014}. The remaining dynamics term $\mathcal{L}_{\mathrm{dyn}}$, which couples the state posterior with the Bayesian GP drift, is detailed in Section~\ref{sec:method_objective}.

\subsection{Irregular-time-aware state posterior}
\label{sec:method_state_posterior}

We now specify the variational smoothing process $\mathbb{Q}_{\phi}(\mathrm{d}\mathbf{X}\vert\mathcal{D})$ in Eq.~\eqref{eq:joint_variational_gpsde}. Following \cite{bartosh2025sde}, we do not parameterize the entire path law directly. Instead, we first specify its one-time marginals $q_{\phi}(\mathbf{x}_t\vert\mathcal{D})$ for arbitrary $t\in[0,T]$, and
then construct a compatible posterior SDE whose path law has these marginals (see Section~\ref{sec:method_objective}).

The key issue is therefore how to condition these marginals on irregularly sampled observations. In SDE Matching, the observation sequence is encoded without its actual timestamps, while the resulting context states are associated with uniformly spaced pseudo-times \cite{bartosh2025sde}. This is appropriate for densely and regularly sampled sequences, but does not preserve unequal observation gaps. We address this limitation by incorporating the true observation times at both the encoding and continuous-time query stages.

Specifically, given
$\mathcal{D}=\{(t_n,\mathbf{y}_n)\}_{n=1}^{N}$, let $t_0=0$ and define $\Delta t_n=t_n-t_{n-1}, n=1,\ldots,N.$
A time-aware recurrent encoder $E_{\phi}$, implemented as a GRU \cite{cho2014learning}, processes each observation together with its elapsed time,
\begin{equation}
\mathbf{h}_{1:N},\mathbf{h}_{\mathrm{g}}
=
E_{\phi}
\!\left(
\{[\mathbf{y}_n,\Delta t_n]\}_{n=1}^{N}
\right),
\label{eq:irregular_encoder}
\end{equation}
where $\mathbf{h}_n$ is the representation of the $n$-th observation and $\mathbf{h}_{\mathrm{g}}$ summarizes the complete observation sequence. By explicitly encoding $\Delta t_n$, the encoder retains the unequal observation gaps ignored by the uniformly spaced pseudo-time representation used in SDE Matching \cite{bartosh2025sde}. For densely and regularly sampled data, where $\Delta t_n$ is constant and the observation index determines the timestamp, the two representations become essentially equivalent.

Because the encoder produces representations only at the observed times, whereas $q_{\phi}(\mathbf{x}_t\vert\mathcal{D})$ must be available at arbitrary continuous times, we construct a query-dependent context using the temporal distances between the query time $t$ and the observations. Specifically, we define temporal weights \cite{shukla2021multi}
\begin{equation}
\alpha_n(t)
=
\frac{
\exp\!\left[-(t-t_n)^2/\ell^2\right]
}{
\sum_{j=1}^{N}
\exp\!\left[-(t-t_j)^2/\ell^2\right]
},
\qquad n=1,\ldots,N,
\label{eq:irregular_time_weights}
\end{equation}
where $\ell>0$ is a temporal bandwidth hyperparameter. The resulting context is
$
\mathbf{c}_{\phi}(t,\mathcal{D})
=
\mathbf{h}_{\mathrm{g}}
+
\sum_{n=1}^{N}\alpha_n(t)\mathbf{h}_n.
$
Thus, the context depends on the true temporal proximity of the observations to $t$, rather than their sequence indices. Together with the elapsed-time encoding in Eq.~\eqref{eq:irregular_encoder}, this makes the state posterior explicitly aware of irregular observation gaps.

Following \cite{bartosh2025sde}, a feed-forward network $H_{\phi}$ maps $[\mathbf{c}_{\phi}(t,\mathcal{D}),t]$ to the mean and log-standard deviation $(\mathbf{m}_{\phi},\log\mathbf{s}_{\phi})$ of a diagonal Gaussian marginal $q_{\phi}(\mathbf{x}_t\vert\mathcal{D})$. Using $\boldsymbol{\epsilon}\sim\mathcal{N}(\mathbf{0},\mathbf{I})$, samples can be reparameterized as
\begin{equation}
\mathbf{x}_t = \mathbf{m}_{\phi}(t,\mathcal{D}) + \mathbf{s}_{\phi}(t,\mathcal{D}) \odot\boldsymbol{\epsilon}.
\label{eq:method_gaussian_state_map}
\end{equation}
Thus, the posterior marginal can be evaluated directly at any $t\in[0,T]$ while accounting for the actual observation times and gaps, without introducing an auxiliary regular inference grid.
Using these marginals, we estimate $\mathcal{L}_{\mathrm{obs}}$ by sampling
$n\sim\mathrm{Unif}\{1,\ldots,N\}$ and
$\boldsymbol{\epsilon}_{\mathrm{obs}}\sim\mathcal{N}(\mathbf{0},\mathbf{I})$.
With
$\mathbf{x}_{t_n}
=
\mathbf{m}_{\phi}(t_n,\mathcal{D})
+
\mathbf{s}_{\phi}(t_n,\mathcal{D})
\odot\boldsymbol{\epsilon}_{\mathrm{obs}}$,
the resulting estimator is
$\widehat{\mathcal{L}}_{\mathrm{obs}}
=
-N\log p(\mathbf{y}_n\vert\mathbf{x}_{t_n})$.

\begin{table*}[t]
\centering
\caption{System identification results under different missing-data ratios.
Results are reported as mean $\pm$ standard deviation over five runs.
Best and second-best results within each observation regime are shown in \textbf{bold} and \underline{underlined}, respectively.
\vspace{-.05in}
}
\label{tab:system_identification}

\small
\renewcommand{\arraystretch}{1.12}
\setlength{\tabcolsep}{3.0pt}

\begin{adjustbox}{max width=\textwidth}
\begin{tabular}{
@{}
ll
c
@{\hspace{7pt}}
c
@{\hspace{7pt}}
>{\columncolor{oursgray}}c
>{\columncolor{oursgray}}c
>{\columncolor{oursgray}}c
@{\hspace{7pt}}
ccc
@{\hspace{7pt}}
>{\columncolor{oursgray}}c
>{\columncolor{oursgray}}c
>{\columncolor{oursgray}}c
@{}
}
\toprule

\multirow{2}{*}{\textsc{Dataset}}
&
\multirow{2}{*}{\textsc{Metric}}
&
\multicolumn{1}{c}{\textsc{GPSSM} \cite{lin2023ensemble}}
&
\multicolumn{1}{c}{\textsc{SDE Matching} \cite{bartosh2025sde}}
&
\multicolumn{3}{c}{\textsc{VGP-SDE} \cite{duncker2019learning}}
&
\multicolumn{3}{c}{\textsc{SING} \cite{hu2026sing}}
&
\multicolumn{3}{c}{\textsc{GP-SDE Matching} \textbf{(Ours)}}
\\[-1pt]

\cmidrule(lr){3-3}
\cmidrule(lr){4-4}
\cmidrule(lr){5-7}
\cmidrule(lr){8-10}
\cmidrule(lr){11-13}

&
&
Dense
&
Dense
&
Dense & 30\% & 60\%
&
Dense & 30\% & 60\%
&
Dense &
30\% &
60\%
\\

\midrule

\multirow{2}{*}{Actuator}
&
\textsc{RMSE} ($\downarrow$)
&
$\mathbf{0.83 \pm 0.05}$
&
$4.30 \pm 9.20$
&
$1.65 \pm 0.01$
&
$1.67 \pm 0.02$
&
$\underline{1.71 \pm 0.05}$
&
$1.10 \pm 1.13$
&
$\underline{1.56 \pm 1.87}$
&
$2.01 \pm 0.71$
&
$\underline{0.85 \pm 0.05}$
&
$\mathbf{0.87 \pm 0.10}$
&
$\mathbf{0.96 \pm 0.09}$
\\

&
\textsc{NLL} ($\downarrow$)
&
$67.24 \pm 8.65$
&
$\underline{29.14 \pm 4.91}$
&
$83.37 \pm 2.69$
&
$93.75 \pm 19.13$
&
$103.14 \pm 19.85$
&
$39.52 \pm 56.0$
&
$\underline{47.51 \pm 64.2}$
&
$\underline{24.64 \pm 20.9}$
&
$\mathbf{1.24 \pm 0.11}$
&
$\mathbf{2.71 \pm 1.15}$
&
$\mathbf{5.21 \pm 2.07}$
\\

\addlinespace[4pt]

\multirow{2}{*}{Ballbeam}
&
\textsc{RMSE} ($\downarrow$)
&
$0.069 \pm 0.011$
&
$0.067 \pm 0.006$
&
$\mathbf{0.055 \pm 0.001}$
&
$\mathbf{0.059 \pm 0.003}$
&
$\mathbf{0.063 \pm 0.003}$
&
$0.212 \pm 0.172$
&
$0.176 \pm 0.135$
&
$0.189 \pm 0.181$
&
$\underline{0.062 \pm 0.015}$
&
$\underline{0.070 \pm 0.018}$
&
$\underline{0.081 \pm 0.026}$
\\

&
\textsc{NLL} ($\downarrow$)
&
$241 \pm 82$
&
$20.565 \pm 26.4$
&
$\mathbf{11.005 \pm 2.2}$
&
$\mathbf{23.637 \pm 12.7}$
&
$\mathbf{37.405 \pm 21.3}$
&
$662 \pm 1290$
&
$1114 \pm 1764$
&
$1098 \pm 1381$
&
$\underline{19.786 \pm 29.1}$
&
$\underline{34.379 \pm 47.1}$
&
$\underline{69.702 \pm 69.4}$
\\

\addlinespace[4pt]

\multirow{2}{*}{Drive}
&
\textsc{RMSE} ($\downarrow$)
&
$0.81 \pm 0.06$
&
$\mathbf{0.74 \pm 0.01}$
&
$0.99 \pm 0.20$
&
$1.35 \pm 0.68$
&
$2.06 \pm 1.42$
&
$0.79 \pm 0.04$
&
$\mathbf{0.78 \pm 0.04}$
&
$\mathbf{0.78 \pm 0.04}$
&
$\underline{0.76 \pm 0.00}$
&
$\underline{0.84 \pm 0.11}$
&
$\underline{0.86 \pm 0.16}$
\\

&
\textsc{NLL} ($\downarrow$)
&
$277 \pm 40$
&
$13.19 \pm 2.70$
&
$9.07 \pm 9.21$
&
$7.10 \pm 6.25$
&
$16.27 \pm 18.10$
&
$\underline{1.86 \pm 0.55}$
&
$\mathbf{1.58 \pm 0.38}$
&
$\mathbf{1.38 \pm 0.17}$
&
$\mathbf{1.79 \pm 0.59}$
&
$\underline{6.73 \pm 7.58}$
&
$\underline{11.92 \pm 14.96}$
\\

\addlinespace[4pt]

\multirow{2}{*}{Dryer}
&
\textsc{RMSE} ($\downarrow$)
&
$\mathbf{0.15 \pm 0.02}$
&
$0.54 \pm 0.02$
&
$1.06 \pm 0.02$
&
$1.04 \pm 0.02$
&
$1.05 \pm 0.03$
&
$0.55 \pm 0.33$
&
$\underline{0.53 \pm 0.23}$
&
$\underline{0.71 \pm 0.21}$
&
$\underline{0.24 \pm 0.03}$
&
$\mathbf{0.26 \pm 0.03}$
&
$\mathbf{0.32 \pm 0.04}$
\\

&
\textsc{NLL} ($\downarrow$)
&
$\underline{4.07 \pm 1.27}$
&
$9.46 \pm 1.36$
&
$74.85 \pm 8.0$
&
$62.30 \pm 19.9$
&
$63.66 \pm 37.6$
&
$50.20 \pm 69.9$
&
$\underline{24.54 \pm 53.2}$
&
$\underline{40.18 \pm 53.6}$
&
$\mathbf{0.09 \pm 0.31}$
&
$\mathbf{0.13 \pm 0.12}$
&
$\mathbf{0.62 \pm 0.42}$
\\

\addlinespace[4pt]

\multirow{2}{*}{GasFurnace}
&
\textsc{RMSE} ($\downarrow$)
&
$\mathbf{1.49 \pm 0.22}$
&
$1.93 \pm 0.02$
&
$3.17 \pm 0.10$
&
$3.38 \pm 0.70$
&
$2.92 \pm 0.42$
&
$2.36 \pm 0.28$
&
$\underline{2.37 \pm 0.26}$
&
$\underline{2.55 \pm 0.38}$
&
$\underline{1.90 \pm 0.04}$
&
$\mathbf{1.88 \pm 0.02}$
&
$\mathbf{1.89 \pm 0.06}$
\\

&
\textsc{NLL} ($\downarrow$)
&
$38.95 \pm 11.74$
&
$26.98 \pm 1.72$
&
$\mathbf{2.93 \pm 0.28}$
&
$\mathbf{3.21 \pm 0.75}$
&
$\mathbf{2.82 \pm 0.60}$
&
$\underline{7.94 \pm 2.11}$
&
$\underline{7.97 \pm 3.17}$
&
$\underline{8.49 \pm 2.20}$
&
$19.59 \pm 4.01$
&
$19.87 \pm 6.70$
&
$17.35 \pm 2.84$
\\

\bottomrule
\end{tabular}
\end{adjustbox}
\vspace{-.05in}
\end{table*}
\begin{algorithm}[t]
\caption{Training GP-SDE Matching}
\label{alg:gpsde_matching}
\begin{algorithmic}[1]
\Require Observations $\mathcal{D}$, inducing inputs $\mathbf{Z}$, variational parameters $(\lambda,\phi)$, and model parameters.
\While{not converged}
    \State Encode
    $\{[\mathbf{y}_n,\Delta t_n]\}_{n=1}^{N}$
    using $E_\phi$ to obtain
    $(\mathbf{h}_{1:N},\mathbf{h}_{\mathrm g})$.

    \State Compute the analytic terms
    $\mathcal{L}_u$ and $\mathcal{L}_0$.

    \State Sample $n\sim\mathrm{Unif}\{1,\ldots,N\}$ and
    $\boldsymbol{\epsilon}_{\mathrm{obs}}\sim\mathcal{N}(\mathbf{0},\mathbf{I})$.
    \State Sample $\mathbf{x}_{t_n} \sim q_{\phi}(\mathbf{x}_{t_n}\vert\mathcal{D})$ and evaluate $\widehat{\mathcal{L}}_{\mathrm{obs}}$.

    \State Sample $t\sim\rho(t)$ and
    $\boldsymbol{\epsilon}\sim\mathcal{N}(\mathbf{0},\mathbf{I})$.
    \State Query
    $(\mathbf{m}_\phi,\mathbf{s}_\phi)$ at $t$ and form
    $\mathbf{x}_t
    =
    \mathbf{m}_\phi
    +
    \mathbf{s}_\phi\odot\boldsymbol{\epsilon}$.

    \State Obtain $(\dot{\mathbf{m}}_\phi,\dot{\mathbf{s}}_\phi)$ and compute $\mathbf{a}_\phi(\mathbf{x}_t,t,\mathcal{D})$.

    \State Compute $\widehat{\mathcal{L}}_{\mathrm{dyn}}$
    using Eq.~\eqref{eq:mc_gp_dynamics_loss}.

    \State Form $\widehat{\mathcal{L}}
    =
    \mathcal{L}_u+\mathcal{L}_0
    +\widehat{\mathcal{L}}_{\mathrm{obs}}
    +\widehat{\mathcal{L}}_{\mathrm{dyn}},$ and update all parameters.
\EndWhile
\end{algorithmic}
\end{algorithm}

\subsection{Simulation-free GP-SDE Matching}
\label{sec:method_objective}

We next construct a path law consistent with the specified state marginals. 
Holding $(\boldsymbol{\epsilon},\mathcal{D})$ fixed, differentiating Eq.~\eqref{eq:method_gaussian_state_map} with respect to time $t$ gives the probability-flow velocity
\begin{equation}
\overline{\mathbf{a}}_{\phi}(\mathbf{x}_t,t,\mathcal{D})
=
\dot{\mathbf{m}}_{\phi}(t,\mathcal{D})
+
\dot{\mathbf{s}}_{\phi}(t,\mathcal{D})
\odot\boldsymbol{\epsilon},
\label{eq:probability_flow_velocity}
\end{equation}
where the dots denote derivatives with respect to $t$. Therefore, a compatible posterior SDE is \cite{song2021scorebased}
\begin{equation}
\mathrm{d}\mathbf{x}_t = \mathbf{a}_{\phi}(\mathbf{x}_t,t,\mathcal{D})\,\mathrm{d}t + \mathbf{G}\,\mathrm{d}\mathbf{W}_t,
\label{eq:posterior_sde}
\end{equation}
with drift $\mathbf{a}_{\phi} = \overline{\mathbf{a}}_{\phi} + \frac{1}{2}\mathbf{A} \nabla_{\mathbf{x}_t} \log q_{\phi}(\mathbf{x}_t\vert\mathcal{D}).$
For the diagonal Gaussian marginal, the score evaluated at the reparameterized sample in Eq.~\eqref{eq:method_gaussian_state_map} is $$
\nabla_{\mathbf{x}_t} \log q_{\phi}(\mathbf{x}_t\vert\mathcal{D})
= - \operatorname{diag}\!\left(
\mathbf{s}_{\phi}^{-1}(t,\mathcal{D})
\right)
\boldsymbol{\epsilon}.
$$
All required time derivatives are obtained by automatic differentiation \cite{paszke2019pytorch}. Note that the SDE in Eq.~\eqref{eq:posterior_sde} has the prescribed state marginals $q_{\phi}(\mathbf{x}_t\vert\mathcal{D})$ and defines the variational path law $\mathbb{Q}_{\phi}(\mathrm{d}\mathbf{X}\vert\mathcal{D})$ in Eq.~\eqref{eq:joint_variational_gpsde}.

Having specified $\mathbb{Q}_{\phi}$, we now turn to the remaining dynamics term $\mathcal{L}_{\mathrm{dyn}}$ in the ELBO. It measures the discrepancy between the variational path law $\mathbb{Q}_{\phi}$ and the generative SDE path law $\mathbb{P}_{\mathbf f}$ in Eq.~\eqref{eq:gpsde_joint_model}, averaged over the variational GP posterior $q_{\lambda}(\mathbf f)$.
Since the two processes share the same diffusion matrix $\mathbf{G}$, Girsanov's theorem expresses this path-space KL through their drift discrepancy \cite{oksendal2003stochastic,archambeau2007gaussian}. After separating the initial-state contribution $\mathcal{L}_0$, we obtain
$$
\mathcal{L}_{\mathrm{dyn}}
=
\frac{1}{2}
\int_{0}^{T}
\mathbb{E}_{q_{\phi}(\mathbf{x}_t\mid\mathcal{D})}
\mathbb{E}_{q_{\lambda}(\mathbf{f})}
\left[
\left\|
\mathbf{a}_{\phi}(\mathbf{x}_t,t,\mathcal{D})
-
\mathbf{f}(\mathbf{x}_t)
\right\|_{\mathbf{A}^{-1}}^{2}
\right]
\,\mathrm{d}t,
$$
where
$\|\mathbf{z}\|_{\mathbf{A}^{-1}}^{2}
=
\mathbf{z}^{\top}\mathbf{A}^{-1}\mathbf{z}$.
Crucially, the GP expectation can be evaluated analytically. Defining $\boldsymbol{\Delta}_{\phi,\lambda}(\mathbf{x},t)
=
\mathbf{a}_{\phi}(\mathbf{x},t)
-
\boldsymbol{\mu}_{\lambda}(\mathbf{x}),$
we obtain
$$
\mathcal{L}_{\mathrm{dyn}}
\!=\!
\frac{1}{2}
\int_{0}^{T}
\mathbb{E}_{q_{\phi}(\mathbf{x}_t\mid\mathcal{D})}
\Big[
\|
\boldsymbol{\Delta}_{\phi,\lambda}
(\mathbf{x}_t,t)
\|_{\mathbf{A}^{\!-\!1}}^{2}
+
\operatorname{tr}\!\left(
\mathbf{A}^{\!-\!1}
\boldsymbol{\Sigma}_{\lambda}(\mathbf{x}_t)
\right)
\Big] \mathrm{d}t .
$$
The first term matches the drift implied by the state posterior to the GP posterior mean $\bm{\mu}_{\lambda}$, while the trace term accounts for the remaining posterior uncertainty $\bm{\Sigma}_{\lambda}$ in the GP drift (see Eq.~\eqref{eq:VSGP}). Without the trace term, the objective would reduce to matching against the GP posterior mean alone.

The remaining time integral is estimated by Monte Carlo using
$t\!\sim\!\rho(t)$ and $\boldsymbol{\epsilon}\!\sim\!\mathcal{N}(\mathbf{0},\mathbf{I})$, where $\rho(t)\!=\!1/T$ is the uniform density over $[0,T]$ in our experiments.
Using the reparameterized state sample in Eq.~\eqref{eq:method_gaussian_state_map}, the resulting estimator is
\begin{equation}
\widehat{\mathcal{L}}_{\mathrm{dyn}}
=
\frac{1}{2\rho(t)}
\left[
\|
\boldsymbol{\Delta}_{\phi,\lambda}
(\mathbf{x}_t,t)
\|_{\mathbf{A}^{-1}}^{2}
+
\operatorname{tr}\!\left(
\mathbf{A}^{-1}
\boldsymbol{\Sigma}_{\lambda}(\mathbf{x}_t)
\right)
\right].
\label{eq:mc_gp_dynamics_loss}
\end{equation}
Thus, the dynamics loss can be evaluated from a single sampled time point and a reparameterized state sample, without simulating an SDE trajectory or sampling a GP function.

The entire GP-SDE Matching algorithm is summarized in Algorithm~\ref{alg:gpsde_matching}. Let $M$ denote the number of inducing points and $H$ the recurrent hidden dimension in the GRU. With a constant number of sampled time points per update, the overall training complexity per iteration is $\mathcal{O}\!\left(NH^2 + DM^3\right)$ where the two terms arise from encoding the $N$ observations and performing sparse GP inference for the $D$ drift dimensions, respectively. Importantly, the complexity is independent of any auxiliary temporal discretization or
SDE solver grid \cite{duncker2019learning}.

\section{Experiments}
\label{sec:experiment}

\subsection{Drift learning and state reconstruction}
\label{subsec:drift_learning_state_reconstruction}
We first evaluate our method on a stochastic Lorenz--63 system,
\[
d\mathbf{x}_t
=
\begin{bmatrix}
\sigma(x_{2,t}-x_{1,t})\\
x_{1,t}(\rho-x_{3,t})-x_{2,t}\\
x_{1,t}x_{2,t}-\beta x_{3,t}
\end{bmatrix}dt
+
\begin{bmatrix}
0.15&0&0\\
0&0.15&0\\
0&0&0.15
\end{bmatrix}
d\mathbf{W}_t,
\]
for $t\in[0,1]$, with $(\sigma,\rho,\beta)=(10,28,8/3)$.
Each trajectory is simulated at 41 equally spaced time points using Euler–Maruyama with a fine integration step, and independent Gaussian observation noise with standard deviation \(0.01\) is added. We use 1,024 trajectories for training and 256 for testing, and standardize all states using training-set statistics.

\begin{figure}
    \centering
    \includegraphics[width=1.03\linewidth]{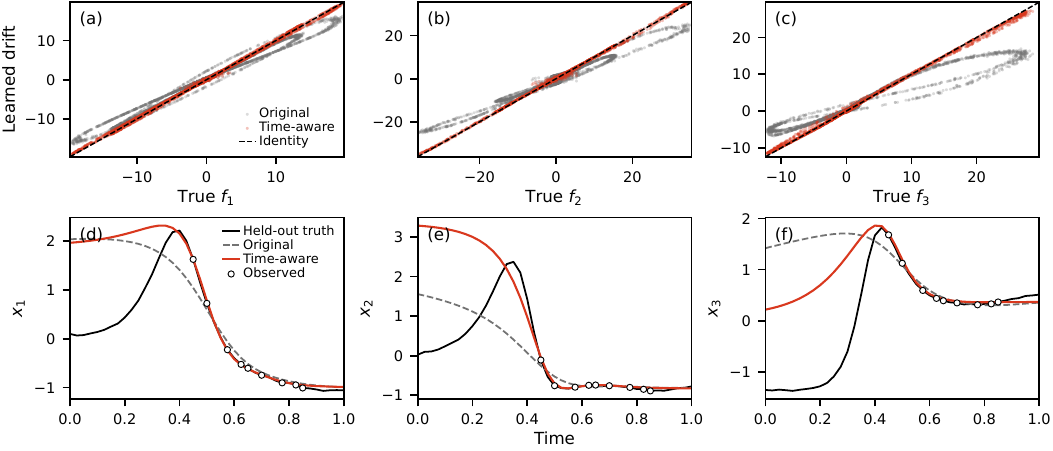}
    \caption{
    Lorenz--63 results under random-window observations. 
    \textbf{Top}: learned versus true drift for the three drift components.
    \textbf{Bottom}: reconstruction of one test trajectory.
    The irregular-time-aware encoder improves both drift learning and state reconstruction.
    \vspace{-.1in}
    }
    \label{fig:Lorenz63}
\end{figure}

For each trajectory, only 9 observations within a randomly located time window of length \(0.4\) are retained (see Fig.~\ref{fig:Lorenz63}), while observations outside the window are hidden. The generative model uses a three-dimensional latent state, a known identity observation model, and fixed diffusion, so only the drift is learned. 
Within GP-SDE Matching, we compare the original SDE Matching encoder with our irregular-time-aware encoder.
Both use a 100-dimensional GRU and the same posterior affine head; the bandwidth in Eq.~\eqref{eq:irregular_time_weights} is fixed at $\ell\!=\!1/9$.

Models are evaluated at unobserved times, shown in Fig.~\ref{fig:Lorenz63}. Under random-window observations, the time-aware encoder recovers the Lorenz drift substantially more accurately than the original encoder (drift RMSE: 0.32 vs. 2.83) and yields markedly better state reconstruction. The improvement is most pronounced in the transient unobserved regions, where the original encoder produces temporally misaligned posterior estimates. These results demonstrate the benefit of explicitly incorporating elapsed times and the actual observation timestamps into the variational posterior.

\vspace{-.1in}
\subsection{Time series forecasting}
\label{subsec:real_data_forcasting}

We next evaluate GP-SDE Matching on five standard system identification datasets (see \cite{doerr2018probabilistic} for more details) to assess predictive performance and robustness to sparse observations. 
For each dataset, the first half is used for training while the second half is reserved for open-loop prediction. Inputs and outputs are normalized using statistics from the training half. To create sparse and irregular training observations, we randomly remove 30\% or 60\% of the training outputs while retaining their timestamps; the dense setting uses all observations. We report raw-scale RMSE and predictive negative log-likelihood (NLL) averaged over five runs.


We compare against four representative baselines: 1) a state-of-the-art discrete-time  \textsc{GPSSM} \cite{lin2023ensemble}; 2) SDE Matching \cite{bartosh2025sde}; and two GP-SDE methods, 3) the variational GP-SDE of Duncker \textit{et al.} (\textsc{VGP-SDE}) \cite{duncker2019learning} and 4) the recent \textsc{SING} \cite{hu2026sing}. 
Since the standard \textsc{GPSSM} and SDE Matching implementations are designed for regularly sampled sequences, we report them only in the dense setting. For all methods, the latent dimension is set to four, and all GP-based methods use 32 inducing points. A linear observation model with Gaussian noise of standard deviation $0.05$ is used throughout.
We use the publicly available implementations and follow their recommended settings whenever applicable.

Table~\ref{tab:system_identification} demonstrates the robustness of GP-SDE Matching across different observation regimes. Among the GP-SDE methods, our approach achieves the lowest dense-data RMSE on four of the five datasets and remains either the best or second-best in RMSE for every dataset at all missing-data ratios. 
In particular, it consistently performs best on Actuator, Dryer, and GasFurnace as the observations become increasingly sparse. 
Predictive NLL is more dataset-dependent. Our method is consistently strongest on Actuator and Dryer and remains competitive on Drive and Ballbeam, whereas \textsc{VGP-SDE} attains lower NLL on Ballbeam and GasFurnace despite substantially larger RMSE on the latter.
Compared with SDE Matching in the dense setting, where the main modeling distinction is the use of a Bayesian GP drift rather than a deterministic neural drift, GP-SDE Matching achieves lower RMSE on four of the five datasets and lower predictive NLL on all five. These results suggest that Bayesian GP drift modeling can improve probabilistic prediction while also enhancing point accuracy in most cases.
Compared with the discrete-time \textsc{GPSSM}, our method has lower RMSE on two of the five datasets but consistently lower predictive NLL on all five. 

Note that GP-SDE Matching is computationally attractive because its training objective avoids latent path simulation. Although \textsc{VGP-SDE}, \textsc{SING}, and \textsc{GPSSM} use different inference mechanisms, they all require computation over an entire latent trajectory or time-discretized state sequence. In contrast, our dynamics objective is evaluated using only pointwise state-posterior and sparse GP quantities, without constructing a latent trajectory or numerical solver grid. As a representative runtime comparison in the dense setting, \textsc{GPSSM} requires 263.1 seconds on average across the five datasets, while the GP-SDE baseline, \textsc{SING}, requires 701.8 seconds on average when using a fine SDE discretization step of $0.0015$. In comparison, GP-SDE Matching requires only 149.9 seconds. Together with the predictive results under 30\% and 60\% missing observations, these runtimes demonstrate the favorable efficiency of the proposed simulation-free formulation.

\vspace{-.1in}
\section{Conclusion}
\label{sec:conclusion}
\vspace{-.05in}
We present GP-SDE Matching, a simulation-free variational framework that extends SDE Matching to Bayesian GP drifts. By analytically marginalizing the sparse GP posterior, we obtain a tractable drift-matching objective that accounts for both the posterior mean and uncertainty of the unknown drift.
We further introduce an irregular-time-aware state posterior for sparse and irregular observations.
Experiments on the Lorenz--63 system and five system-identification benchmarks demonstrate improved drift recovery and state reconstruction under irregular observations, competitive forecasting across observation regimes, and favorable training efficiency. 

\clearpage
\balance
\bibliographystyle{IEEEbib}
\bibliography{refs}

\end{document}